\documentclass[letterpaper]{article} % DO NOT CHANGE THIS
\usepackage{aaai2026}  % DO NOT CHANGE THIS
\usepackage{times}  % DO NOT CHANGE THIS
\usepackage{helvet}  % DO NOT CHANGE THIS
\usepackage{courier}  % DO NOT CHANGE THIS
\usepackage[hyphens]{url}  % DO NOT CHANGE THIS
\usepackage{graphicx} % DO NOT CHANGE THIS
\usepackage{natbib}  % DO NOT CHANGE THIS AND DO NOT ADD ANY OPTIONS TO IT
\usepackage{caption} % DO NOT CHANGE THIS AND DO NOT ADD ANY OPTIONS TO IT
\usepackage{algorithm}
\usepackage{algorithmic}

\usepackage{booktabs}
\usepackage{multirow}
\usepackage{multicol}
\usepackage{amsmath}
\usepackage{subcaption}
\usepackage{colortbl}
\usepackage{xcolor}
\nocopyright
\usepackage{newfloat}
\usepackage{listings}
\DeclareCaptionStyle{ruled}{labelfont=normalfont,labelsep=colon,strut=off} % DO NOT CHANGE THIS
\floatstyle{ruled}
\newfloat{listing}{tb}{lst}{}
\floatname{listing}{Listing}
\title{From Text to Simulation: A Multi-Agent LLM Workflow for\\ Automated Chemical Process Design}
\author{
    Xufei Tian\textsuperscript{\rm 1},
    Wenli Du\textsuperscript{\rm 1,2}\thanks{Corresponding author.},
    Shaoyi Yang\textsuperscript{\rm 1},
    Han Hu\textsuperscript{\rm 1},
    Hui Xin\textsuperscript{\rm 1},
    Shifeng Qu\textsuperscript{\rm 1},
    Ke Ye\textsuperscript{\rm 1}
}
\affiliations{
    \textsuperscript{\rm 1}State Key Laboratory of Industrial Control Technology, East China University of Science and Technology\\
    \textsuperscript{\rm 2}Huzhou Institute of Industrial Control Technology, Huzhou, China
    
    \{xufei.tian, sy.yang, han.hu, xinhui, shifengqu, yeke\}@mail.ecust.edu.cn, 
    
    wldu@ecust.edu.cn
}

\begin{document}

\maketitle

\begin{figure*}[h]
\centering
\includegraphics[width=\textwidth]{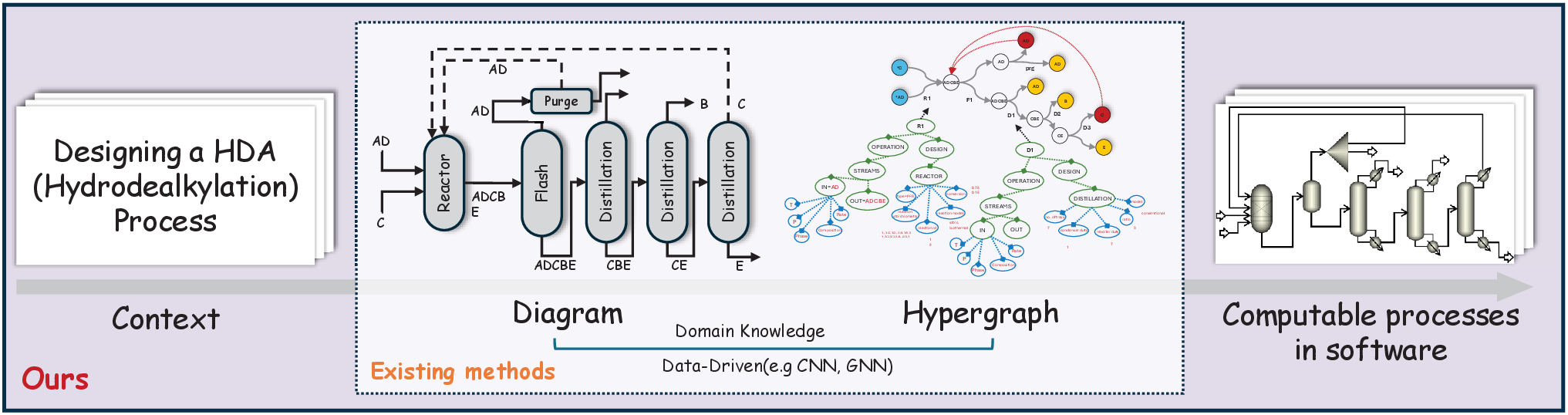}
\caption{Representation of different stages of chemical process design workflow. ``Diagram'' refers to a general process flow diagram\cite{douglas1988conceptual}, while ``Hypergraph'' represents a process hypergraph with parameter annotations \cite{mann2024esfiles}. The entire stage requires a lot of manual operation design, while existing automated methods only focus on various representations of graphs. Our method achieves cross-dimensional and fully automated implementation
}
\label{fig:figure1}
\end{figure*}

\begin{abstract}
Process simulation is a critical cornerstone of chemical engineering design. Current automated chemical design methodologies focus mainly on various representations of process flow diagrams. However, transforming these diagrams into executable simulation flowsheets remains a time-consuming and labor-intensive endeavor, requiring extensive manual parameter configuration within simulation software. In this work, we propose a novel multi-agent workflow that leverages the semantic understanding capabilities of large language models(LLMs) and enables iterative interactions with chemical process simulation software, achieving end-to-end automated simulation from textual process specifications to computationally validated software configurations for design enhancement. Our approach integrates four specialized agents responsible for task understanding, topology generation, parameter configuration, and evaluation analysis, respectively, coupled with Enhanced Monte Carlo Tree Search to accurately interpret semantics and robustly generate configurations. Evaluated on Simona, a large-scale process description dataset, our method achieves a 31.1\% improvement in the simulation convergence rate compared to state-of-the-art baselines and reduces the design time by 89. 0\% compared to the expert manual design. This work demonstrates the potential of AI-assisted chemical process design, which bridges the gap between conceptual design and practical implementation. Our workflow is applicable to diverse process-oriented industries, including pharmaceuticals, petrochemicals, food processing, and manufacturing, offering a generalizable solution for automated process design.
\end{abstract}


\section{Introduction}
The design of chemical processes forms the foundation of numerous industrial sectors, including petrochemicals, pharmaceuticals, sustainable energy, and material manufacturing. Despite decades of development in process simulation software, translating high-level process requirements into executable simulation models remains a significant challenge in chemical engineering workflows\cite{mann2023group}. Engineers typically spend weeks transforming conceptual designs into detailed configurations, manually specifying hundreds of interdependent parameters. This manual configuration process not only consumes valuable engineering time, but also limits exploration of innovative design alternatives.

Current approaches to chemical process design automation have achieved progress in specific aspects, yet they fail to provide comprehensive solutions. As illustrated in the middle section of Figure~\ref{fig:figure1}, existing methods typically focus on intermediate representations, either generating process flow diagrams or constructing graph-based representations \cite{balhorn2022flowsheet, d2006process}. These approaches terminate at the structural representation level, leaving a critical gap between abstract process descriptions and executable simulations. While flowsheet-based methods capture the visual topology of processes, and hypergraph representations encode connectivity relationships, neither can directly generate simulation-ready configurations executable in industrial software. This limitation forces engineers to manually bridge the gap between these representations and functional process models, perpetuating bottlenecks in design workflows. The fundamental challenge extends beyond simply representing the process structure to generating complete executable configurations that satisfy complex engineering constraints. Process simulation requires a precise specification of thermodynamic models, operating conditions, equipment parameters, and control strategies, details that cannot be adequately captured by structural representations alone. Moreover, the interdependencies among these parameters create a complex optimization space where changes in one unit operation cascade throughout the entire process. Existing methods focus on either topology generation or parameter optimization in isolation, failing to address this holistic challenge, resulting in designs that appear structurally sound but prove infeasible or suboptimal when implemented in simulation software.

To address these limitations, we propose a novel approach, as shown in Figure~\ref{fig:figure1}, which directly transforms natural language process descriptions into executable computational processes for simulation software. Unlike existing methods that terminate at intermediate representations, we achieve an end-to-end workflow from textual specifications to validated simulation configurations. Our work leverages large language models (LLMs) within a multi-agent workflow to interpret process requirements, construct appropriate topologies, configure detailed parameters, and validate designs through direct integration with professional simulation software. This approach eliminates the manual transformation steps required by current methods, enabling rapid iteration and exploration of design alternatives.

The key innovation of our approach lies in bridging the representation-to-implementation gap through three integrated components. First, we employ a multi-agent system that decomposes complex design tasks into specialized subtasks: task understanding, topology generation, parameter configuration, and evaluation analysis. Each agent leverages LLM capabilities while incorporating domain-specific constraints and engineering principles. Second, we introduce an Enhanced Monte Carlo Tree Search (E-MCTS) algorithm that efficiently explores the design space while maintaining feasibility constraints throughout the search process. Third, and most critically, we establish seamless bidirectional communication with industrial simulation software, enabling real-time validation and iterative refinement of generated configurations. This integration ensures that our outputs are not merely theoretical constructs but practically executable designs validated through rigorous computational verification.
In summary, our contributions are as follows.
\begin{itemize}
    \item We propose the first end-to-end workflow that directly generates executable chemical process simulations from natural language descriptions, bypassing the limitations of intermediate representations employed by existing methods.
    
    \item We develop a multi-agent architecture that effectively combines LLMs' semantic understanding capabilities with chemical engineering domain knowledge, achieving accurate interpretation and implementation of complex process requirements.
    
    \item Through extensive experiments, we demonstrate that our work achieves an 80.3\% success rate in generating valid executable process configurations, reducing design time by 89.0\% compared to manual methods while maintaining industrial grade quality.
    
    \item We establish a new paradigm for AI-assisted chemical engineering that transcends structural representations to deliver complete, simulatable solutions, opening new possibilities for accelerating process design innovation.
\end{itemize}

\section{Related Work}
\subsection{Chemical Process Simulation}
The traditional paradigm of chemical process simulation follows a multistage workflow: beginning with thermodynamic feasibility analysis, followed by process flow diagram (PFD) construction within feasible domains, and culminating in validation through specialized simulation software until convergence\cite{mann2022hybrid, venkatasubramanian2019promise, tula2017computer}. To accelerate this design cycle, researchers have explored digitization and symbolic representation approaches for flowsheets\cite{weininger1988smiles, zhang2019pattern, tula2019hybrid}. Early work employed traditional machine learning methods for automatic flowsheet completion\cite{vogel2023sfiles, theisen2023digitization}, while eSFILES\cite{mann2024esfiles} integrated domain knowledge with data-driven approaches to enable flowsheet analysis and weighting. Recent studies have incorporated large language models to enhance the intelligence of process synthesis\cite{vogel2023learning}. The fundamental limitation of PFD representations lies in their level of abstraction: while they effectively capture unit operations and connectivity relationships, they omit critical configuration parameters essential for simulation convergence. Traditionally, setting parameters is a process that is not only time-consuming but also highly dependent on expert knowledge.

Existing automation approaches face three core challenges when addressing complex chemical processes: (i) lack of deep integration with specialized process simulation software, preventing rapid validation of generated design solutions; (ii) absence of simulation-based feedback mechanisms, precluding iterative optimization of configuration parameters; (iii) difficulty in bridging the semantic gap between high-level process descriptions and executable simulation configurations. This leaves numerous process designs at the conceptual level, with validation of large-scale processes proving particularly challenging.

\subsection{Agentic Workflow}
The remarkable perception and reasoning capabilities of LLM-driven agents in complex environments have spurred extensive exploration and rapid advancement in multi-agent workflows\cite{song2023llm, zhao2024expel, xu2023lemur, liu2023bolaa, ramesh2021zero, yao2023react, wang2022self}. Current multi-agent workflow research spans both general-purpose and domain-specific applications\cite{pang2025paper2poster, zhuge2024agent, chen2025re, liu2024agentbench, liu2024llava}, including automated workflow generation\cite{wu2021flow, qiao2024benchmarking, li2024autoflow}, code generation\cite{cen2025sqlfixagent, qian2023chatdev, xu2023lemur} and the collaboration of multi-agent\cite{zhang2024g, chen2023agentverse, li2023camel, zhuge2024gptswarm}. While ChemCrow\cite{bran2023chemcrow} and Coscientist\cite{boiko2023autonomous} demonstrate the potential of LLMs in chemistry, a substantial gap persists between these approaches and industrial-grade applications: (i) insufficient consideration of engineering constraints and scale-up effects; (ii) inability to establish effective feedback loops with specialized engineering software. This gap motivates our focus on applying multi-agent workflows to the more challenging domain of industrial engineering implementation. Our approach processes textual descriptions to iteratively generate computationally verifiable chemical process configurations or thermodynamic analysis requests via multi-agent workflows, presenting significant challenges in configuration evaluation at each iteration.

\section{Method}
\subsection{Overview}
\begin{figure*}[t]
\centering
\includegraphics[width=\textwidth]{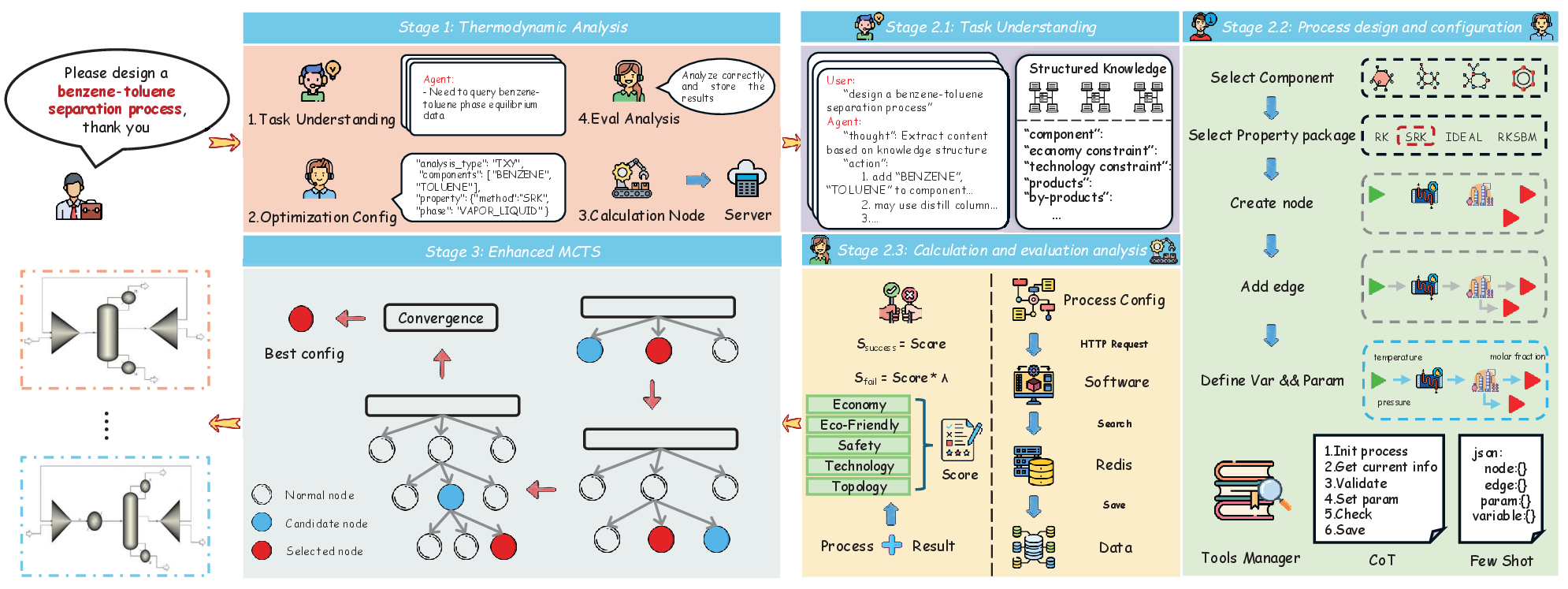}
\caption{The overview of our comprehensive workflow, consisting of four key agents: (i) the \textit{Task Understanding Agent}, which parses user inputs to extract key parameters, constraints, and design objectives. (ii) the \textit{Topology Generation Agent}, which determines appropriate chemical components and thermodynamic methods, selects suitable unit operations and establishes their connections. (iii) the \textit{Parameter Configuration Agent}, which assigns operating conditions to each unit within the generated topology, such as temperature and pressure. (iv) the \textit{Evaluation Analysis Agent}, which assesses simulation results to determine whether to iterate or terminate.}
\label{fig:overview}
\end{figure*}

We propose a multi-agent collaborative workflow built upon LangGraph for automating chemical process design from natural language descriptions. The workflow transforms textual process specifications into executable simulation configurations through coordinated operation of specialized agents, employing E-MCTS for systematic exploration of the design space. Our system accepts natural language inputs ranging from high-level process requirements (e.g., ``design an ethylene cracking process'') to detailed specifications encompassing equipment constraints, operating parameters, and economic or environmental objectives. It generates complete configuration files directly executable in professional chemical process simulation software, containing fully specified process topologies, operating conditions, and convergence results. As illustrated in Figure~\ref{fig:overview}, our workflow coordinates four core agents: (1) \textit{Task Understanding Agent}, (2) \textit{Topology Generation Agent}, (3) \textit{Parameter Configuration Agent},  (4) \textit{Evaluation Analysis Agent}.

Our work consists of a main workflow accompanied by a subordinate workflow: the aforementioned main design workflow and an auxiliary thermodynamic analysis workflow executed prior to process design. This preliminary analysis validates critical component properties and phase equilibrium behavior, establishing a robust foundation for the main workflow and preventing fundamental errors in subsequent iterations. The iterative refinement mechanism employs Enhanced Monte Carlo Tree Search to guide exploration, systematically navigating the design space while maintaining computational efficiency to ensure high-quality solutions within reasonable timeframes. This closed-loop optimization emulates human expert iterative refinement while leveraging computational advantages of automated exploration.

\subsection{Task Understanding Agent}
The Task Understanding Agent serves as the entry of our multi-agent workflow, designed to parse ambiguous user descriptions into structured requirements for downstream processing. Users typically provide natural language descriptions of their process design ideas, which often lack critical specifications necessary for chemical process configuration. For instance, when a user queries ``Please help me check whether two substances form an azeotrope'', they are essentially requesting thermodynamic temperature properties of a binary system, though this technical requirement remains implicit in their description.

The fundamental challenge addressed by this agent is the semantic gap between informal user expressions and the rigorous specifications required by process simulation tools. To bridge this gap, we have developed a comprehensive data structure that encodes critical attribute information aligned with our simulation requirements. The extracted information is then structured into a standardized format that serves as the communication protocol between agents. This structured representation ensures that subsequent agents receive unambiguous specifications while preserving the flexibility to handle diverse user queries. Thus, the task understanding agent enables robust downstream processing and reduces the likelihood of configuration errors in the final simulation setup.

\subsection{Topology Generation Agent}
The Topology Generation Agent transforms process description into chemical process topologies, bridging the gap between abstract user requirements and executable flowsheets. This agent achieves automated conversion from functional requirements to structural design by leveraging domain knowledge and systematic design principles, and is the only agent that does not participate in the thermodynamic analysis configuration workflow. The core design philosophy of this agent is to decouple topology generation from parameter configuration. Considering that the final output of the multi-agent workflow is a JSON configuration file verifiable by process simulation software, and that LLM-based agents struggle to directly generate configurations that strictly meet computational requirements, we have designed a specialized tool function library. This enables the agent to focus on high-level decisions regarding unit selection and connectivity relationships, thereby significantly improving design efficiency. The agent employs a multi-candidate strategy to generate several feasible alternatives, a feature that manifests in the subsequent E-MCTS. Each topology is represented as a directed graph $G=(V,E)$, where the node set $V$ corresponds to unit operations and the edge set E represents material flow connections. Each design is accompanied by a unique process manager reference identifier, enabling downstream agents to access and modify the topology structure to support the iterative optimization process.

The Topology Generation Agent accesses process design tools through a globally unified WorkflowToolsManager, implementing the following core functions: (i) Process initialization, creating independent process manager instances for each candidate solution to ensure parallel exploration of the design space; (ii) Adding components and selecting property methods; (iii) Unit operation instantiation, incorporating various equipment such as reactors, distillation columns, and heat exchangers; (iv) Stream connections, establishing material and energy flow relationships between unit operations; (v) Design persistence, saving completed topologies to a tree structure for subsequent processing. During subsequent iterations, this agent also supports cascading deletion of unit nodes and edge removal operations to facilitate topology design modifications.

\subsection{Parameter Configuration Agent}
The Parameter Configuration Agent leverages the powerful contextual understanding and reasoning capabilities of LLMs, integrating chemical engineering domain knowledge to transform topological structures into executable chemical processes by determining optimal operating parameters for each unit operation. This agent accomplishes two key functions: (i) Assigning initial operating conditions guided by a model template library; (ii) Performing further parameter adjustments based on simulation feedback. To enhance the rationality of parameter configurations, we employ chain-of-thought\cite{wei2022chain} prompting strategies to guide LLMs through step-by-step reasoning. Additionally, we have compiled extensive high-quality parameter configuration examples from feasible processes, utilizing few-shot learning to improve the quality of LLM-generated parameter configurations.

The agent passes adjusted configurations to dedicated computational nodes, interfacing with external simulation tools through asynchronous HTTP protocols, enabling rigorous validation via chemical process simulation software while maintaining a closed-loop optimization mechanism. This integration is crucial, as LLM-based agents, while excelling at high-level reasoning, struggle to directly generate parameters that strictly satisfy the complex interdependencies of chemical processes.

\subsection{Evaluation Analysis Agent}
The Evaluation Analysis agent serves as a critical quality assurance component in our workflow, fulfilling dual responsibilities: assessing current process configurations and providing actionable improvement directions for subsequent iterations. Following validation through external simulation software via computational nodes, the agent performs quantitative scoring of both the process configuration and simulation results. When simulations fail to converge, a penalty factor $\lambda \in (0,1)$  is applied. 
The agent employs a comprehensive multi-dimensional evaluation framework that assesses process designs across the following key dimensions:

\textbf{Economic Feasibility} ($\mathcal{E}_f$) : This dimension evaluates the economic feasibility and cost-effectiveness of process configurations, encompassing capital expenditure, operating costs, and return on investment to ensure the process design maintains economic competitiveness.

\textbf{Environmental Sustainability} ($\mathcal{E}_s$) : This metric assesses environmental impact and sustainability, covering pollutant emissions, carbon footprint, and resource recycling to advance green chemistry objectives.

\textbf{Process Safety} ($\mathcal{P}_s$) : This criterion ensures inherent process safety through evaluation of hazardous material handling, safety system configurations, and accident risk assessment to identify potential hazards.

\textbf{Technical Feasibility} ($\mathcal{T}_f$) : This dimension validates the maturity and implementability of process technologies, evaluating reaction mechanism rationality, separation technology applicability, and operational control complexity.

\textbf{Topological Rationality} ($\mathcal{T}_r$) : This metric evaluates the completeness and optimization degree of process topology, including unit operation connectivity logic and material flow rationality to verify efficient process structure design.
The overall evaluation score for each iteration is computed as a weighted combination of the individual dimensions.
\begin{equation}
\mathcal{S}_i = w_1 \cdot \mathcal{E}_f + w_2 \cdot \mathcal{E}_s + w_3 \cdot \mathcal{P}_s + w_4 \cdot \mathcal{T}_f + w_5 \cdot \mathcal{T}_r
\end{equation}
where $\mathcal{S}_i$ represents the comprehensive evaluation score for the $i$-th iteration, and the weights for the five evaluation dimensions are 0.35, 0.25, 0.15, 0.15, and 0.10\cite{seider2016product}, respectively, prioritizing economic and environmental considerations in industrial applications. In case of simulation failure, a penalty factor is applied:
$\mathcal{S}_i^{fail} = \lambda \cdot \mathcal{S}_i$, where $\lambda$ is set to 0.3.

\subsection{Enhanced MCTS}
Inspired by AFlow\cite{zhang2024aflow}, our Enhanced Monte Carlo Tree Search(E-MCTS) approach aims to address a critical limitation in iterative search processes, where configurations that fail to converge are often overlooked despite containing valuable design insights worthy of further exploration. In E-MCTS structure, each tree node represents a complete process configuration rather than individual unit operations, enabling the exploration of diverse complex chemical process designs.

\textbf{Initialization} E-MCTS begins with three initial nodes generated by workflow, derived from the user's input, serving as the foundation for systematic exploration. The root node is created as a virtual node, with the three initial configurations becoming its child nodes. We initialize the dynamic candidate pool as an empty set and establish key parameters, including the dynamic weight $\alpha(0)$ for initial exploration focus and the exploration coefficient $c(0)$.

\textbf{Selection with Dual-Layer Value Evaluation} We propose a dual-layer value evaluation system to address the unique challenge where failed configurations may contain valuable design insights. Our approach distinguishes between immediate value (derived directly from simulation scores) and potential value (considering multi-dimensional metrics). For failed configurations, we recognize exceptional performance in specific dimensions as indicators of refinement potential. The combined value integrates both perspectives through dynamic weighting:
\begin{equation}
\begin{aligned}
    V(n_i, t) &= \alpha(t) \cdot V_{\text{imm}}(n_i) + (1-\alpha(t)) \cdot V_{\text{pot}}(n_i)
\end{aligned}
\end{equation}
where $\alpha(t)$ transitions from exploration-focused to exploitation-focused as the search progresses. We further enhance node selection by incorporating domain-specific features, augmenting the standard UCB formula:
\begin{equation}
\text{UCB}_{\text{enh}}(n_i, t) = V_{\text{comb}}(n_i, t) + c(t)\sqrt{\frac{\ln v_p}{v_i}} + \Psi(n_i)
\end{equation}
where the feature weighting function $\Psi(n_i)$ incorporates recent improvement indicators, variance-based sensitivity measures, and depth preferences.

\textbf{Expansion} During the expansion phase, each selected node generates three child configurations. We execute a new workflow iteration to update configurations, leveraging both the selected node's configuration and historical insights from past iterations. The experience encompasses all modifications with their corresponding improvements or failures, along with precise logs of predictions and expected outputs, maximizing insight extraction from previous attempts.

\textbf{Dynamic Revisit Mechanism} 
In contrast to traditional MCTS which abandons low-scoring branches, we maintain a dynamic candidate pool containing the second and third highest-scoring nodes that have not been revisited. When global improvement stagnates, the algorithm selects nodes for revisitation based on the gap between potential and realized performance:
\begin{equation}
n_{\text{rev}} = \arg\max_{n \in \mathcal{R}} \left(V_{\text{pot}}(n) - V_{\text{imm}}(n)\right)
\end{equation}
This mechanism prevents premature convergence and enables the algorithm to escape local optima by revisiting promising configurations that initially failed due to suboptimal parameters.

\textbf{Terminal Conditions} We implement comprehensive termination criteria tailored to chemical process optimization: (1) discovery of configurations exceeding target performance, (2) reaching iteration limits, (3) global improvement stagnation across all active leaves, or (4) high convergence indicated by concentrated visit frequencies and stable scores. The final solution selection combines solution quality (70\% weight), search confidence as measured by visit frequency (20\% weight), and stability indicators (10\% weight), ensuring robust and reliable process configurations.

This E-MCTS algorithm effectively addresses the challenges of chemical process optimization by recognizing value in failed configurations, dynamically balancing exploration and exploitation, and implementing intelligent revisit strategies to systematically navigate the vast design space.

\begin{table*}[!h]
\centering
\begin{tabular}{lcccccc|cc}
\toprule
\multirow{2}{*}{Methods} & \multicolumn{6}{c|}{Main Scores} & \multicolumn{2}{c}{Efficiency}\\
\cmidrule(lr){2-7} \cmidrule(lr){8-9}
& $\mathcal{E}_f \uparrow$ & $\mathcal{E}_s\uparrow $ & $\mathcal{P}_s \uparrow$ & $\mathcal{T}_f \uparrow$ & $\mathcal{T}_r \uparrow$ & $\mathcal{S} \uparrow $ & Time(s) $\downarrow$ & SCR(\%)$\uparrow$ \\
\midrule
GPT-4o\cite{achiam2023gpt} & 73.6 & 77.2 & 71.4 & 75.5 & 69.8 & 34.36 & \textbf{86.13} & 23.4 \\
Claude-Sonnet-4\cite{claude-4} & 74.3 & 73.9 & 76.1 & 72.7 & 71.5 & 33.16 & 89.24 & 21.2 \\
\midrule
Swarm\cite{swarm} & 79.9 & 79.4 & 80.3 & 79.7 & 80.1 & 49.32 & 149.21 & 45.4 \\
AutoGen\cite{wu2024autogen} & 79.4 & 78.2 & 80.1 & 79.6 & 78.3 & 48.27 & 303.16 & 44.3 \\
CrewAI\cite{crewai} & 79.8 & 77.5 & 78.1 & 79.3 & 79.8 & 49.95 & 389.09 & 47.6 \\
MetaGPT\cite{hong2023metagpt} & 81.2 & 79.9 & 80.5 & 81.0 & 80.4 & 51.98 & 612.33 & 49.2 \\
\midrule
Expert manual & \textbf{87.2} & 85.6 & \textbf{84.6} & 86.4 & \textbf{85.9} & \textbf{86.16} & 8301.91 & \textbf{100.0}\\
\midrule
Ours-4o & 85.7 & 84.9 & 83.8 & 86.4 & 84.2 & 73.01 & 892.04 & 79.6 \\
Ours*-4o & 85.8 & \textbf{86.0} & 84.3 & \textbf{86.5} & 85.5 & 73.88 & 913.21 & 80.3 \\
\bottomrule
\end{tabular}
\caption{Evaluation of our workflow on Simona dataset. ``Ours'' represents a pure process configuration generation workflow, while ``Ours*'' represents a configuration generation workflow that introduces thermodynamic analysis workflow}
\label{tab:performance}
\end{table*}

\section{Experiment}

\subsection{Baselines}
We evaluate our approach against four categories of baselines: (1) End-to-end methods: We prompt GPT-4o and Claude Sonnet 4 to directly generate JSON request configurations. (2) Multi-agent workflows: We employ Swarm, AutoGen, CrewAI, MetaGPT to benchmark our multi-agent workflow's effectiveness. (3) Human expert baseline: We invite three chemical engineering experts to perform manual design and computational validation. (4) Our methods: We evaluate two variants of our proposed approach—a base workflow without thermodynamic analysis workflow and an enhanced version incorporating thermodynamic analysis.

\subsection{Dataset}
Our experiments are conducted on a self-constructed dataset called Simona, consisting of 1000 process descriptions carefully designed by chemical engineering experts. The dataset encompasses process descriptions of varying complexity, primarily distributed across: (1) Unit operations of different difficulty levels (i.e., nodes in a process flowsheet), including mixers, distillation columns, among others. (2) Process descriptions with varying levels of detail—for relatively simple cases, we provide comprehensive process descriptions including component specifications, unit model specifications, and partial parameter specifications; for more challenging cases, we deliberately provide ambiguous process descriptions with certain critical information withheld.

\subsection{Implementation Details}
Our workflow is built upon LangGraph\cite{langgraph} (v0.4.10) with LangChain (v0.3.26) providing the foundation for LLM interactions, as our workflow involves iterative loops that align well with LangGraph's characteristics. We employ gpt-4o as the base language model for all agents. The temperature parameters for the task understanding agent and evaluation analysis agent are set to 0.1 to ensure relatively stable task comprehension and result evaluation. In contrast, the temperature parameters for the topology generation agent and parameter configuration agent are set to 0.9 to promote greater diversity and innovation in the process schemes generated across different branches. For chemical process simulation, we integrate with our self-developed chemical process simulation software through HTTP APIs.

\subsection{Main Results}
\textbf{Main Scores.} Table~\ref{tab:performance}(left) presents the evaluation results across five dimensions along with the overall score, which combines scores from both successful and failed simulation cases. Due to our conservative scoring mechanism in the evaluation agent, our method does not demonstrate substantial advantages across individual dimensions compared to baseline approaches. However, our method achieves significantly higher overall scores than other end-to-end and general multi-agent methods, primarily attributed to our simulation convergence rate in process simulation software. Notably, expert manual design exhibits clear superiority in economic feasibility, as evidenced by the optimized topology structures and computational results. Our method achieves comparable performance to expert design across other dimensions and ranks second in overall scores. Since scoring alone cannot fully capture the significance of automated design workflows, we further analyze efficiency metrics.

\textbf{Efficiency.} Table~\ref{tab:performance}(right) presents efficiency metrics across different methods. We recruited three chemical engineering PhD students to design processes based on requirement descriptions and recorded their average design time. Regarding Simulation Convergence Rate (SCR)—defined as the ratio of successfully converged process designs to total requirements—expert manual design achieves 100\% SCR due to extensive parameter optimization experience. Our method achieves 80.3\% SCR, substantially outperforming other baseline methods. In terms of time efficiency, GPT-4o demonstrates remarkable speed, generating process configurations in merely 86.13 seconds on average. However, its SCR of only 23.4\% indicates that it can merely handle simple processes with detailed parameter specifications, offering limited practical value for automated process design. While expert manual design achieves exceptional SCR, it requires considerable time investment—8301.91 seconds on average—with complex processes often demanding significantly longer durations. Our method maintains high SCR while achieving 89.0\% reduction in design time compared to manual approaches, representing a significant advancement in accelerating chemical process design.

\begin{figure}[h]
\centering
\includegraphics[width=8cm]{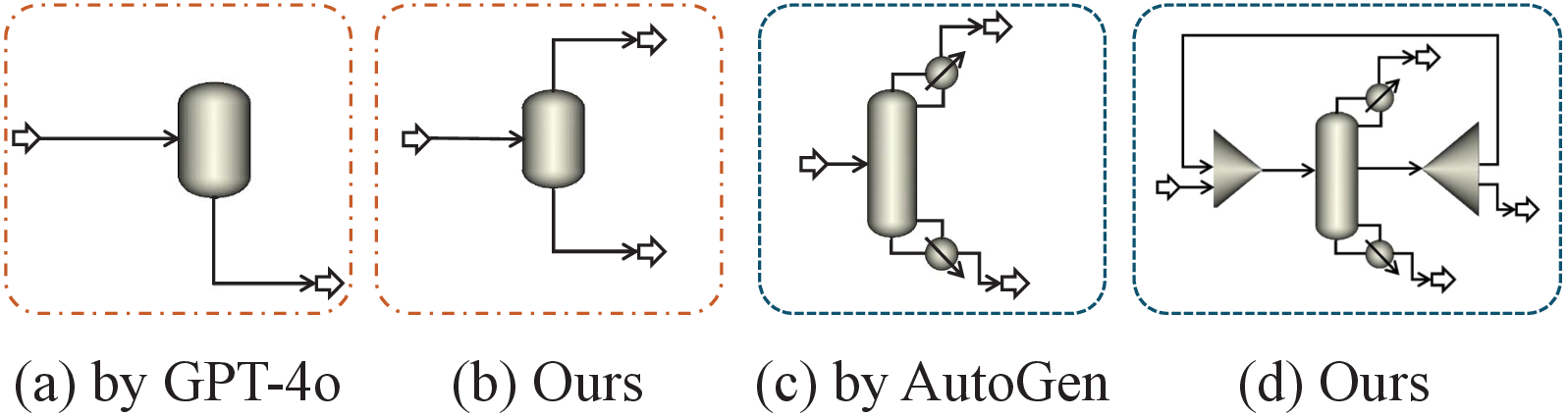}
\caption{Illustration of the design generated by different methods, where different colors represent different cases.}
\label{fig:qualitative}
\end{figure}

\textbf{Qualitative Analysis.} To intuitively compare output differences across methods, we conduct qualitative analysis as shown in Figure~\ref{fig:qualitative}. We select representative configuration outputs from different baselines and visualized the corresponding process flowsheets using chemical process simulation software. In the flash drum separation process, the output of GPT-4o exhibits a fundamental flaw: while a flash drum requires one inlet stream and two outlet streams (liquid and vapor phases), the generated configuration includes only a single outlet stream. This represents a common failure resulting from the absence of computational validation checks. For the benzene-toluene separation process, the flowsheet generated by AutoGen meets basic functional requirements but lacks off-specification product recycle loops. Our work, accounting for practical economic constraints, incorporates reflux streams to ensure product quality specifications. While the overall process topologies appear similar, the economic performance differs dramatically, demonstrating the value of our chemical process design agent workflow.

\subsection{Ablation Study}
\textbf{Analysis of Key Components.}
We conducted ablation experiments on two critical components: Task Understanding Agent(TUA) and E-MCTS, with results presented in Table~\ref{tab:TUA}. Upon removing the Task Understanding Agent, the workflow output scores decreased significantly. This indicates that without systematic context extraction and understanding of user inputs, relying solely on the process design and parameter optimization agents to directly process raw user inputs leads to critical constraint omissions, resulting in degraded generation quality and substantially reduced iteration efficiency. Similarly, disabling E-MCTS resulted in a significant drop in the convergence rate. Our experiments revealed that initial topology and parameter configuration flaws severely impair subsequent iteration efficiency, causing the process to terminate prematurely when score improvements plateau.

\begin{table}[h]
\centering
\setlength{\tabcolsep}{1mm}
\begin{tabular}{lccccccc}  % l=左对齐, c=居中, r=右对齐
\toprule
\textbf{Model} & $\mathcal{E}_f \uparrow$ & $\mathcal{E}_s\uparrow $ & $\mathcal{P}_s \uparrow$ & $\mathcal{T}_f \uparrow$ & $\mathcal{T}_r \uparrow$ & Iter$\downarrow$ & SCR$\uparrow$ \\
\midrule
Ours*-4o & \textbf{85.8} & \textbf{86.0} & \textbf{84.3} & \textbf{86.5} & \textbf{85.5} & 15 & \textbf{80.3} \\
w/o E-MCTS & 84.9 & 85.5 & 82.9 & 84.9 & 84.2 & \textbf{7} & 74.8 \\
w/o TUA & 83.2 & 83.1 & 81.0 & 83.3 & 82.6 & 18 & 78.9 \\
\bottomrule
\end{tabular}
\caption{Ablation study of the impact of key components on execution efficiency}
\label{tab:TUA}
\end{table}

\textbf{Impact of E-MCTS Nodes.}
We performed ablation experiments on the number of child nodes in E-MCT, evaluating settings of 2, 3, 4, and 5 children per expansion. As shown in Table~\ref{tab:MCTSNode}, lower child node counts accelerate convergence and reduce token consumption in Simona, yet yield lower SCR due to limited exploration, which constrains the generation of high-quality solutions. Conversely, higher child node counts slow convergence and increase token consumption, but the expanded search space enhances the probability of discovering superior configurations, resulting in improved SCR. To balance convergence efficiency, token costs, and solution quality, we adopted a child node count of 3 as the optimal trade-off.

\begin{table}[h]
\centering
\setlength{\tabcolsep}{4pt}
\begin{tabular}{cccc}  % l=左对齐, c=居中, r=右对齐
\toprule
\textbf{E-MCTS Nodes} & \textbf{Time}(s) $\downarrow$ & \textbf{Tokens}(K) $\downarrow$ & \textbf{SCR}(\%) $\uparrow$ \\
\midrule
2 & \textbf{736.55} & \textbf{794.2} & 79.8\\
3 & 913.21 & 994.7 & 80.3\\
4 & 1102.72 & 1192.1 & 80.5\\
5 & 1197.09 & 1325.6 & \textbf{80.6}\\
\bottomrule
\end{tabular}
\caption{The impact of the number of E-MCTS nodes on time and token efficiency}
\label{tab:MCTSNode}
\end{table}

\textbf{Analysis of ICL Methods.}
Table~\ref{tab:ICL} presents ablation studies on in-context learning (ICL) within our workflow. We removed the process-construction chain-of-thought (CoT) prompting from the topology generation agent and few-shot parameter examples from the parameter configuration agent. Experimental results demonstrate that these components are crucial: Their absence compromises both process design rationality and computational convergence, as initial parameter values significantly impact computational efficiency in complex chemical processes.

\begin{table}[h]
\centering
\setlength{\tabcolsep}{1mm}
\begin{tabular}{lcccccc}  % l=左对齐, c=居中, r=右对齐
\toprule
\textbf{Model} & $\mathcal{E}_f \uparrow$ & $\mathcal{E}_s\uparrow $ & $\mathcal{P}_s \uparrow$ & $\mathcal{T}_f \uparrow$ & $\mathcal{T}_r \uparrow$ & SCR($\%$)$\uparrow$\\
\midrule
Ours*-4o & \textbf{85.8} & \textbf{86.0} & \textbf{84.3} & \textbf{86.5} & \textbf{85.5} & \textbf{80.3} \\
w/o CoT & 84.6 & 84.1 & 84.0 & 85.2 & 84.9 & 78.8 \\
w/o Few-Shot & 85.3 & 85.7 & 83.9 & 85.8 & 85.2 & 77.9 \\
\bottomrule
\end{tabular}
\caption{Ablation study of ICL methods on execution efficiency}
\label{tab:ICL}
\end{table}

\section{Conclusion}
In this paper, we have presented a comprehensive study on automated chemical process design and validation that accelerates both the design phase and the simulation-based verification phase, significantly reducing the design time for chemical engineering practitioners. We have introduced a novel automated workflow for generating computational configurations in chemical process simulation software, departing from the traditional paradigm of manual configuration from flowsheets. This establishes a foundational infrastructure for accelerating iterative design validation in chemical process engineering. Our work decomposes the process design validation workflow and leverages E-MCTS to improve design robustness. Experiments on Simona dataset demonstrate the effectiveness of our approach. Compared to existing methods, our work not only generates feasible topological structures but also produces outputs that are computationally verifiable. Furthermore, our research underscores the potential of automated chemical process design. While there remains scope for further refinement compared to expert-designed solutions, the significant advantages in time efficiency establish a practical and extensible methodology for future chemical process design endeavors.

\section{Acknowledgments}
This work was supported by the National Key Research and Development Program of China (2022YFB3305900), National Natural Science Foundation of China (62394345, 62103151), the Programme of Introducing Talents of Discipline to Universities (the 111 Project) under Grant B17017 and Fundamental Research Funds for the Central Universities (222202517006). 

\bibliography{aaai2026}

@inproceedings{wu2024autogen,
  title={Autogen: Enabling next-gen LLM applications via multi-agent conversations},
  author={Wu, Qingyun and Bansal, Gagan and Zhang, Jieyu and Wu, Yiran and Li, Beibin and Zhu, Erkang and Jiang, Li and Zhang, Xiaoyun and Zhang, Shaokun and Liu, Jiale and others},
  booktitle={First Conference on Language Modeling},
  year={2024}
}

@article{zhang2024aflow,
  title={Aflow: Automating agentic workflow generation},
  author={Zhang, Jiayi and Xiang, Jinyu and Yu, Zhaoyang and Teng, Fengwei and Chen, Xionghui and Chen, Jiaqi and Zhuge, Mingchen and Cheng, Xin and Hong, Sirui and Wang, Jinlin and others},
  journal={arXiv preprint arXiv:2410.10762},
  year={2024}
}

@article{li2024autoflow,
  title={Autoflow: Automated workflow generation for large language model agents},
  author={Li, Zelong and Xu, Shuyuan and Mei, Kai and Hua, Wenyue and Rama, Balaji and Raheja, Om and Wang, Hao and Zhu, He and Zhang, Yongfeng},
  journal={arXiv preprint arXiv:2407.12821},
  year={2024}
}

@inproceedings{hong2023metagpt,
  title={MetaGPT: Meta programming for a multi-agent collaborative framework},
  author={Hong, Sirui and Zhuge, Mingchen and Chen, Jonathan and Zheng, Xiawu and Cheng, Yuheng and Wang, Jinlin and Zhang, Ceyao and Wang, Zili and Yau, Steven Ka Shing and Lin, Zijuan and others},
  booktitle={The Twelfth International Conference on Learning Representations},
  year={2023}
}

@article{achiam2023gpt,
  title={Gpt-4 technical report},
  author={Achiam, Josh and Adler, Steven and Agarwal, Sandhini and Ahmad, Lama and Akkaya, Ilge and Aleman, Florencia Leoni and Almeida, Diogo and Altenschmidt, Janko and Altman, Sam and Anadkat, Shyamal and others},
  journal={arXiv preprint arXiv:2303.08774},
  year={2023}
}

@misc{claude-4,
    title={Introducing Claude 4},
    author={Anthropic},
    year={2025},
    howpublished={"\url{https://www.anthropic.com/news/claude-4}"},
    note = " Accessed: 2025-6-25"
}

@misc{langgraph,
    title={LangGraph},
    author={langchain-ai},
    year={2024},
    howpublished={"\url{https://github.com/langchain-ai/langgraph}"},
    note = " Accessed: 2025-6-10"
}

@misc{crewai,
    title={CrewAI},
    author={crewAIInc},
    year={2023},
    howpublished={"\url{https://github.com/crewAIInc/crewAI}"},
    note = " Accessed: 2025-6-10"
}

@misc{swarm,
    title={Swarm},
    author={OpenAI},
    year={2024},
    howpublished={"\url{https://github.com/openai/swarm}"},
    note = " Accessed: 2025-6-10"
}

@article{douglas1988conceptual,
  title={Conceptual design of chemical processes},
  author={Douglas, James Merrill},
  journal={(no Title)},
  year={1988}
}

@article{zhang2019pattern,
  title={Pattern recognition in chemical process flowsheets},
  author={Zhang, Tong and Sahinidis, Nikolaos V and Siirola, Jeffrey J},
  journal={AIChE Journal},
  volume={65},
  number={2},
  pages={592--603},
  year={2019},
  publisher={Wiley Online Library}
}

@article{tula2019hybrid,
  title={Hybrid method and associated tools for synthesis of sustainable process flowsheets},
  author={Tula, Anjan K and Eden, Mario R and Gani, Rafiqul},
  journal={Computers \& Chemical Engineering},
  volume={131},
  pages={106572},
  year={2019},
  publisher={Elsevier}
}

@article{weininger1988smiles,
  title={SMILES, a chemical language and information system. 1. Introduction to methodology and encoding rules},
  author={Weininger, David},
  journal={Journal of chemical information and computer sciences},
  volume={28},
  number={1},
  pages={31--36},
  year={1988},
  publisher={ACS Publications}
}

@book{d2006process,
  title={Process flow sheet generation \& design through a group contribution approach},
  author={d'Anterroches, Lo{\"\i}c},
  year={2006},
  publisher={Technical University of Denmark}
}

@article{vogel2023learning,
  title={Learning from flowsheets: A generative transformer model for autocompletion of flowsheets},
  author={Vogel, Gabriel and Balhorn, Lukas Schulze and Schweidtmann, Artur M},
  journal={Computers \& Chemical Engineering},
  volume={171},
  pages={108162},
  year={2023},
  publisher={Elsevier}
}

@article{mann2023group,
  title={Group contribution-based property modeling for chemical product design: A perspective in the AI era},
  author={Mann, Vipul and Gani, Rafiqul and Venkatasubramanian, Venkat},
  journal={Fluid Phase Equilibria},
  volume={568},
  pages={113734},
  year={2023},
  publisher={Elsevier}
}

@incollection{balhorn2022flowsheet,
  title={Flowsheet recognition using deep convolutional neural networks},
  author={Balhorn, Lukas Schulze and Gao, Qinghe and Goldstein, Dominik and Schweidtmann, Artur M},
  booktitle={Computer Aided Chemical Engineering},
  volume={49},
  pages={1567--1572},
  year={2022},
  publisher={Elsevier}
}

@article{mann2022hybrid,
  title={Hybrid, interpretable machine learning for thermodynamic property estimation using grammar2vec for molecular representation},
  author={Mann, Vipul and Brito, Karoline and Gani, Rafiqul and Venkatasubramanian, Venkat},
  journal={Fluid Phase Equilibria},
  volume={561},
  pages={113531},
  year={2022},
  publisher={Elsevier}
}

@article{venkatasubramanian2019promise,
  title={The promise of artificial intelligence in chemical engineering: Is it here, finally?},
  author={Venkatasubramanian, Venkat},
  journal={AIChE Journal},
  volume={65},
  number={1},
  year={2019}
}

@article{vogel2023sfiles,
  title={SFILES 2.0: an extended text-based flowsheet representation},
  author={Vogel, Gabriel and Hirtreiter, Edwin and Schulze Balhorn, Lukas and Schweidtmann, Artur M},
  journal={Optimization and Engineering},
  volume={24},
  number={4},
  pages={2911--2933},
  year={2023},
  publisher={Springer}
}

@article{mann2024esfiles,
  title={eSFILES: Intelligent process flowsheet synthesis using process knowledge, symbolic AI, and machine learning},
  author={Mann, Vipul and Sales-Cruz, Mauricio and Gani, Rafiqul and Venkatasubramanian, Venkat},
  journal={Computers \& Chemical Engineering},
  volume={181},
  pages={108505},
  year={2024},
  publisher={Elsevier}
}

@article{theisen2023digitization,
  title={Digitization of chemical process flow diagrams using deep convolutional neural networks},
  author={Theisen, Maximilian F and Flores, Kenji Nishizaki and Balhorn, Lukas Schulze and Schweidtmann, Artur M},
  journal={Digital Chemical Engineering},
  volume={6},
  pages={100072},
  year={2023},
  publisher={Elsevier}
}

@article{bran2023chemcrow,
  title={Chemcrow: Augmenting large-language models with chemistry tools},
  author={Bran, Andres M and Cox, Sam and Schilter, Oliver and Baldassari, Carlo and White, Andrew D and Schwaller, Philippe},
  journal={arXiv preprint arXiv:2304.05376},
  year={2023}
}

@article{boiko2023autonomous,
  title={Autonomous chemical research with large language models},
  author={Boiko, Daniil A and MacKnight, Robert and Kline, Ben and Gomes, Gabe},
  journal={Nature},
  volume={624},
  number={7992},
  pages={570--578},
  year={2023},
  publisher={Nature Publishing Group UK London}
}

@article{tula2017computer,
  title={A computer-aided software-tool for sustainable process synthesis-intensification},
  author={Tula, Anjan Kumar and Babi, Deenesh K and Bottlaender, Jack and Eden, Mario R and Gani, Rafiqul},
  journal={Computers \& Chemical Engineering},
  volume={105},
  pages={74--95},
  year={2017},
  publisher={Elsevier}
}

@article{wu2021flow,
  title={Flow: A modular learning framework for mixed autonomy traffic},
  author={Wu, Cathy and Kreidieh, Abdul Rahman and Parvate, Kanaad and Vinitsky, Eugene and Bayen, Alexandre M},
  journal={IEEE Transactions on Robotics},
  volume={38},
  number={2},
  pages={1270--1286},
  year={2021},
  publisher={IEEE}
}

@inproceedings{liu2024llava,
  title={Llava-plus: Learning to use tools for creating multimodal agents},
  author={Liu, Shilong and Cheng, Hao and Liu, Haotian and Zhang, Hao and Li, Feng and Ren, Tianhe and Zou, Xueyan and Yang, Jianwei and Su, Hang and Zhu, Jun and others},
  booktitle={European conference on computer vision},
  pages={126--142},
  year={2024},
  organization={Springer}
}

@article{chen2025re,
  title={Re-Aligning Language to Visual Objects with an Agentic Workflow},
  author={Chen, Yuming and Feng, Jiangyan and Zhang, Haodong and Gong, Lijun and Zhu, Feng and Zhao, Rui and Hou, Qibin and Cheng, Ming-Ming and Song, Yibing},
  journal={arXiv preprint arXiv:2503.23508},
  year={2025}
}

@inproceedings{yao2023react,
  title={React: Synergizing reasoning and acting in language models},
  author={Yao, Shunyu and Zhao, Jeffrey and Yu, Dian and Du, Nan and Shafran, Izhak and Narasimhan, Karthik and Cao, Yuan},
  booktitle={International Conference on Learning Representations (ICLR)},
  year={2023}
}

@article{wang2022self,
  title={Self-consistency improves chain of thought reasoning in language models},
  author={Wang, Xuezhi and Wei, Jason and Schuurmans, Dale and Le, Quoc and Chi, Ed and Narang, Sharan and Chowdhery, Aakanksha and Zhou, Denny},
  journal={arXiv preprint arXiv:2203.11171},
  year={2022}
}

@inproceedings{
liu2024agentbench,
title={AgentBench: Evaluating {LLM}s as Agents},
author={Xiao Liu and Hao Yu and Hanchen Zhang and Yifan Xu and Xuanyu Lei and Hanyu Lai and Yu Gu and Hangliang Ding and Kaiwen Men and Kejuan Yang and Shudan Zhang and Xiang Deng and Aohan Zeng and Zhengxiao Du and Chenhui Zhang and Sheng Shen and Tianjun Zhang and Yu Su and Huan Sun and Minlie Huang and Yuxiao Dong and Jie Tang},
booktitle={The Twelfth International Conference on Learning Representations},
year={2024},
url={https://openreview.net/forum?id=zAdUB0aCTQ}
}

@article{qiao2024benchmarking,
  title={Benchmarking agentic workflow generation},
  author={Qiao, Shuofei and Fang, Runnan and Qiu, Zhisong and Wang, Xiaobin and Zhang, Ningyu and Jiang, Yong and Xie, Pengjun and Huang, Fei and Chen, Huajun},
  journal={arXiv preprint arXiv:2410.07869},
  year={2024}
}

@article{pang2025paper2poster,
  title={Paper2Poster: Towards Multimodal Poster Automation from Scientific Papers},
  author={Pang, Wei and Lin, Kevin Qinghong and Jian, Xiangru and He, Xi and Torr, Philip},
  journal={arXiv preprint arXiv:2505.21497},
  year={2025}
}

@inproceedings{cen2025sqlfixagent,
  title={SQLFixAgent: Towards Semantic-Accurate Text-to-SQL Parsing via Consistency-Enhanced Multi-Agent Collaboration},
  author={Cen, Jipeng and Liu, Jiaxin and Li, Zhixu and Wang, Jingjing},
  booktitle={Proceedings of the AAAI Conference on Artificial Intelligence},
  volume={39},
  number={1},
  pages={49--57},
  year={2025}
}

@article{qian2023chatdev,
  title={Chatdev: Communicative agents for software development},
  author={Qian, Chen and Liu, Wei and Liu, Hongzhang and Chen, Nuo and Dang, Yufan and Li, Jiahao and Yang, Cheng and Chen, Weize and Su, Yusheng and Cong, Xin and others},
  journal={arXiv preprint arXiv:2307.07924},
  year={2023}
}

@article{zhuge2024agent,
  title={Agent-as-a-judge: Evaluate agents with agents},
  author={Zhuge, Mingchen and Zhao, Changsheng and Ashley, Dylan and Wang, Wenyi and Khizbullin, Dmitrii and Xiong, Yunyang and Liu, Zechun and Chang, Ernie and Krishnamoorthi, Raghuraman and Tian, Yuandong and others},
  journal={arXiv preprint arXiv:2410.10934},
  year={2024}
}

@article{wei2022chain,
  title={Chain-of-thought prompting elicits reasoning in large language models},
  author={Wei, Jason and Wang, Xuezhi and Schuurmans, Dale and Bosma, Maarten and Xia, Fei and Chi, Ed and Le, Quoc V and Zhou, Denny and others},
  journal={Advances in neural information processing systems},
  volume={35},
  pages={24824--24837},
  year={2022}
}

@inproceedings{song2023llm,
  title={Llm-planner: Few-shot grounded planning for embodied agents with large language models},
  author={Song, Chan Hee and Wu, Jiaman and Washington, Clayton and Sadler, Brian M and Chao, Wei-Lun and Su, Yu},
  booktitle={Proceedings of the IEEE/CVF international conference on computer vision},
  pages={2998--3009},
  year={2023}
}

@inproceedings{zhao2024expel,
  title={Expel: Llm agents are experiential learners},
  author={Zhao, Andrew and Huang, Daniel and Xu, Quentin and Lin, Matthieu and Liu, Yong-Jin and Huang, Gao},
  booktitle={Proceedings of the AAAI Conference on Artificial Intelligence},
  volume={38},
  number={17},
  pages={19632--19642},
  year={2024}
}

@article{xu2023lemur,
  title={Lemur: Harmonizing natural language and code for language agents},
  author={Xu, Yiheng and Su, Hongjin and Xing, Chen and Mi, Boyu and Liu, Qian and Shi, Weijia and Hui, Binyuan and Zhou, Fan and Liu, Yitao and Xie, Tianbao and others},
  journal={arXiv preprint arXiv:2310.06830},
  year={2023}
}

@article{zhang2024g,
  title={G-designer: Architecting multi-agent communication topologies via graph neural networks},
  author={Zhang, Guibin and Yue, Yanwei and Sun, Xiangguo and Wan, Guancheng and Yu, Miao and Fang, Junfeng and Wang, Kun and Chen, Tianlong and Cheng, Dawei},
  journal={arXiv preprint arXiv:2410.11782},
  year={2024}
}

@article{chen2023agentverse,
  title={Agentverse: Facilitating multi-agent collaboration and exploring emergent behaviors in agents},
  author={Chen, Weize and Su, Yusheng and Zuo, Jingwei and Yang, Cheng and Yuan, Chenfei and Qian, Chen and Chan, Chi-Min and Qin, Yujia and Lu, Yaxi and Xie, Ruobing and others},
  journal={arXiv preprint arXiv:2308.10848},
  volume={2},
  number={4},
  pages={6},
  year={2023}
}

@article{li2023camel,
  title={Camel: Communicative agents for" mind" exploration of large language model society},
  author={Li, Guohao and Hammoud, Hasan and Itani, Hani and Khizbullin, Dmitrii and Ghanem, Bernard},
  journal={Advances in Neural Information Processing Systems},
  volume={36},
  pages={51991--52008},
  year={2023}
}

@article{liu2023bolaa,
  title={Bolaa: Benchmarking and orchestrating llm-augmented autonomous agents},
  author={Liu, Zhiwei and Yao, Weiran and Zhang, Jianguo and Xue, Le and Heinecke, Shelby and Murthy, Rithesh and Feng, Yihao and Chen, Zeyuan and Niebles, Juan Carlos and Arpit, Devansh and others},
  journal={arXiv preprint arXiv:2308.05960},
  year={2023}
}

@inproceedings{ramesh2021zero,
  title={Zero-shot text-to-image generation},
  author={Ramesh, Aditya and Pavlov, Mikhail and Goh, Gabriel and Gray, Scott and Voss, Chelsea and Radford, Alec and Chen, Mark and Sutskever, Ilya},
  booktitle={International conference on machine learning},
  pages={8821--8831},
  year={2021},
  organization={Pmlr}
}

@inproceedings{zhuge2024gptswarm,
  title={Gptswarm: Language agents as optimizable graphs},
  author={Zhuge, Mingchen and Wang, Wenyi and Kirsch, Louis and Faccio, Francesco and Khizbullin, Dmitrii and Schmidhuber, J{\"u}rgen},
  booktitle={Forty-first International Conference on Machine Learning},
  year={2024}
}

@book{seider2016product,
  title={Product and process design principles: synthesis, analysis and evaluation},
  author={Seider, Warren D and Lewin, Daniel R and Seader, JD and Widagdo, Soemantri and Gani, Rafiqul and Ng, Ka Ming},
  year={2016},
  publisher={John Wiley \& Sons}
}

\end{document}